\documentclass[10pt,twocolumn,letterpaper]{article}

\usepackage{iccv}
\usepackage{times}
\usepackage{epsfig}
\usepackage{graphicx}
\usepackage{amsmath}
\usepackage{amssymb}

\usepackage[dvipsnames]{xcolor}
\usepackage{multirow}
\usepackage{makecell}
\usepackage{threeparttable}
\usepackage{booktabs}
\usepackage{stfloats}
\usepackage{float}
\usepackage{tabularx}
\usepackage{afterpage}

\newcommand{\tb}[1]{#1}
\newcommand{\tr}[1]{#1}
\newcommand{\tg}[1]{#1}

\newfloat{algorithm}{tbp}{loa}
\providecommand{\algorithmname}{Algorithm}
\floatname{algorithm}{\protect\algorithmname}

\usepackage[pagebackref=true,breaklinks=true,letterpaper=true,colorlinks,bookmarks=false]{hyperref}

\iccvfinalcopy % *** Uncomment this line for the final submission

\def\iccvPaperID{0} % *** Enter the ICCV Paper ID here

\ificcvfinal\fi

\def\eg{\textit{e.g.,~}}

\def\ie{\textit{i.e.~}}

\def\etal{\textit{et~al.~}}

\newcommand{\RR}{\mathcal R}
\newcommand{\AP}{\text{AP}_{\text{Q}}}
\newcommand{\mAP}{\text{mAP}_{\text{Q}}}
\newcommand{\ROxford}{$\mathcal{R}$Oxford}
\newcommand{\RParis}{$\mathcal{R}$Paris}
\newcommand{\ROM}{$\mathcal{R}$1M}
\newcommand{\indic}{1 \! \! 1}

\begin{document}

\title{Learning with Average Precision: Training \\
	   Image Retrieval with a Listwise Loss}

\author{Jerome Revaud \qquad 
	    Jon Almaz\'an \qquad 
	    Rafael Sampaio de Rezende \qquad 
	    C\'esar Roberto de Souza \\[12pt]
NAVER LABS Europe \\ 
 %\texttt{\small{}{}{}{}\{jerome.revaud,rafael.sampaio-de-rezende,cesar.desouza,jon.almazan\}@naverlabs.com}{\small{}{}{}{} % For a paper whose authors are all at the same institution,
%\and Second Author}\\
% {\small{}{}{}{} Institution2}\\
% {\small{}{}{}{} First line of institution2 address}\\
% {\small{}{}{}{} }\texttt{\small{}{}{}{}secondauthor@i2.org}{\small{}{}{}{} }}
}
\maketitle

\begin{abstract}
Image retrieval can be formulated as a ranking problem where the goal is to order database images by decreasing similarity to the query. Recent deep models for image retrieval have outperformed traditional methods by leveraging ranking-tailored loss functions, but important theoretical and practical problems remain. First, rather than directly optimizing the global ranking, they minimize an upper-bound on the essential loss, which does not necessarily result in an optimal mean average precision (mAP). Second, these methods require significant engineering efforts to work well, \eg special pre-training and hard-negative mining. In this paper we propose instead to directly optimize the global mAP by leveraging recent advances in listwise loss formulations. Using a histogram binning approximation, the AP can be differentiated and thus employed to end-to-end learning. 
Compared to existing losses, the proposed method considers thousands of images simultaneously at each iteration and eliminates the need for ad hoc tricks. It also establishes a new state of the art on many standard retrieval benchmarks. Models and evaluation scripts have been made available at: \url{https://europe.naverlabs.com/Deep-Image-Retrieval/}.
\end{abstract}

\vspace{-4mm}
\section{Introduction\label{sec:intro}}

\vspace{-1mm}
\begin{figure*}[t!]
    \vspace*{-2mm}
    \centering
	\includegraphics[width=0.95\linewidth]{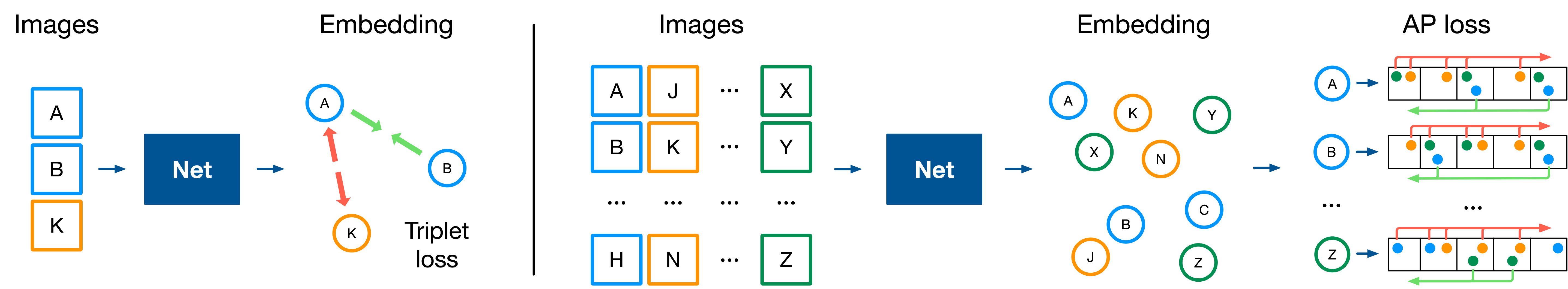}
	\caption{Illustration of the differences between a local ranking loss (here triplet-based) and our listwise loss. 
	The triplet loss (left) performs gradient updates based on a small number of examples, which is not guarranteed to be aligned with 
	a ranking metric. 
	In contrast, the listwise loss (right) considers a large number of images simultaneously and directly optimizes the Average-Precision
	computed from these images. }
	\label{fig:loss}
    \vspace*{-4mm}
\end{figure*}

Image retrieval consists in
finding, given a query, all images containing relevant 
content within a large database. Relevance
here is defined at the instance level and retrieval typically consists in ranking 
in top positions database images with the same object instance as the one in the query.
This important technology serves as a building block for popular applications
such as image-based item identification (\eg fashion 
items~\cite{Corbiere2017,Liu2016,Wang2018} or products~\cite{ProdSearch16})
and automatic organization of personal photos \cite{Guy2018}. 

Most instance retrieval approaches rely on computing image signatures
that are robust to viewpoint variations and other types of noise.
Interestingly, signatures extracted by deep learned models have recently outperformed 
keypoint-based traditional methods \cite{Gordo2016,Gordo2017,Radenovic2018a}. 
This good performance was enabled by the ability of deep models to leverage a family of loss functions 
well-suited to the ranking problem. 
Compared to classification losses previously used for retrieval with less success~\cite{Babenko2015,Babenko2014,Razavian2014}, 
ranking-based loss functions directly optimize for the end task, enforcing intra-class discrimination and 
more fine-grained instance-level image representations~\cite{Gordo2017}.
Ranking losses used to date consider either
image pairs \cite{Radenovic2016}, triplets \cite{Gordo2017}, quadruplets \cite{Chen2017}, or $n$-tuples \cite{Sohn2016}.
Their common principle is to subsample a small set of images, verify that
they locally comply with the ranking objective, perform a small
model update if they do not, and repeat these steps until convergence.

Despite their effectiveness, important theoretical and practical problems remain.
In particular, it has been shown that these ranking losses are upper bounds on a 
quantity known as the \emph{essential loss}~\cite{Liu2009}, which in turn is an 
upper bound on standard retrieval metrics such as mean average precision 
(mAP)~\cite{Liu2011}. Thus, optimizing these ranking losses
is not guaranteed to give results that also optimize mAP.
Hence there is no theoretical guarantee that these approaches
yield good performance in a practical system.
Perhaps for this reason, many tricks are required to obtain good results,
such as pre-training for classification~\cite{Arandjelovic15,Gordo2017}, 
combining multiple losses~\cite{   Chen2018,Chen2017b}, 
and using complex hard-negative mining strategies~\cite{
    Faghri2018,Hardwood2017,Manmatha2017,Mishchuk2017}. 
These engineering heuristics involve additional hyper-parameters and are 
notoriously complicated to implement and tune~\cite{He2018,Ustinova2016}.

In this paper, we investigate a new type of ranking loss that
remedy these issues altogether by directly optimizing mAP (see Fig. \ref{fig:loss}).
Instead of considering a couple of images at a time,
it optimizes the global ranking of thousands of images simultaneously.
This practically renders the aforementioned tricks unnecessary while 
improving the performance at the same time.
Specifically, we leverage recent advances in listwise loss functions that
allow to reformulate AP using histogram binning~\cite{He2018b,He2018,Ustinova2016}. 
AP is normally non-smooth and not differentiable, and cannot be
directly optimized in gradient-based frameworks. 
Nevertheless, histogram binning (or soft-binning) is differentiable and
can be used to replace the non-differentiable sorting operation in the AP,
making it amenable to deep learning. 
He \etal \cite{He2018} recently presented outstanding results 
in the context of patch verification, patch retrieval and image matching based on this technique. 

In this work, we follow the same path and 
present an image retrieval approach directly optimized for mAP.
To that aim, we train with large batches of high-resolution images
that would normally considerably exceed the memory of a GPU. 
We therefore introduce an optimization scheme that renders training feasible for arbitrary batch sizes, image resolutions and network depths.
In summary, we make three main contributions: 

\vspace{-1mm}

\begin{itemize}
\item We present, for the first time, an approach to image retrieval leveraging a listwise
ranking loss directly optimizing the mAP.
It hinges upon a dedicated optimization scheme that handles extremely large batch sizes with arbitrary image resolutions and network depths.
\item We demonstrate the many benefits of using our listwise loss in terms of coding effort, training budget and final performance via a \textit{ceteris paribus} analysis on the loss.
\item We outperform the state-of-the-art results for comparable training sets and
networks. 
\end{itemize}

The paper is organized as follows: 
Section \ref{sec:related} discusses related work, 
Section \ref{sec:method} describes the proposed method, 
Section \ref{sec:xp} present an experimental study, and 
Section \ref{sec:Conclusion} presents our conclusions.

\section{Related work\label{sec:related}}

Early works on instance retrieval relied on local patch descriptors (\eg SIFT \cite{Lowe2004}),
aggregated using bag-of-words representations \cite{Csurka04}
or more elaborate schemes~\cite{Gong2013,Gordo2012,Spyromitros2014,Jegou14},
to obtain image-level signatures that could then be compared to one 
another in order to find their closest matches. 
Recently, image signatures extracted with CNNs have emerged as an alternative.
While initial work used neuron activations extracted
from off-the-shelf networks pre-trained for classification~\cite{Babenko2015,Babenko2014,Radenovic2016,Razavian2014,Tolias16},
it was later shown that networks could be trained specifically for the task
of instance retrieval in an end-to-end manner using a siamese network~\cite{Gordo2017,Radenovic2018a}.
The key was to leverage a loss function that optimizes ranking instead of classification.
This class of approaches represents the current state of the art in image retrieval with global representations~\cite{Gordo2017,Radenovic2018a}. 

Image retrieval can indeed be seen as a learning to rank 
problem \cite{Burges2005,Cao2007,Liu2009,Trotman2005}. In 
this framework, the task is to determine in which (partial) 
order elements from the training set should appear. 
It is solved using metric learning combined with an appropriate 
ranking loss. Most works in image retrieval have considered 
pairwise (\eg contrastive~\cite{Radenovic2016})
or tuplewise (\eg triplet-based~\cite{Gordo2016,Schroff2015}, $n$-tuple-based~\cite{Sohn2016}) 
loss functions, which we call {\em local} loss functions because they act
on a fixed and limited number of examples before computing the gradient.
For such losses, training consists in repeatedly sampling random 
and difficult pairs or triplets of images, computing the loss, 
and backpropagating its gradient. 
However, several works \cite{Faghri2018,He2018,Hermans2017,Movshovitz2017,Rippel2016}
pointed out that properly optimizing a local loss can be a challenging
task, for several reasons. First, it requires a number of ad hoc heuristics such
as pre-training for classification~\cite{Arandjelovic15,Gordo2017}, combining several losses~\cite{
	Chen2018, 
	Chen2017b} 
and biasing the sampling of image pairs by mining hard or semi-hard negative examples 
\cite{Faghri2018,Hardwood2017,Manmatha2017,Mishchuk2017}.
Besides being non trivial~\cite{Harwood2017,Shi2016,Wang2017},
mining hard examples is often time consuming. 
Another major problem that has been overlooked so far 
is the fact that local loss functions only optimize an upper bound
on the true ranking loss~\cite{Liu2009,Ustinova2016}. 
As such, there is no theoretical guarantee that the minimum of the loss actually 
corresponds to the minimum of the true ranking loss. % such as the Average Precision.

In this paper, we take a different approach and directly optimize 
the mean average precision (mAP) metric. While the AP is a non-smooth and
non-differentiable function, He \etal \cite{He2018b,He2018} have recently
shown that it can be approximated based on the differentiable 
approximation to histogram binning proposed in \cite{Ustinova2016}
(and also used in \cite{Cakir2017}).
This approach radically differs from those based on local losses.
The use of histogram approximations to mAP is called {\em listwise} in~\cite{Liu2009}, as the loss function takes a variable 
(possibly large) number of examples at the same time and optimizes 
their ranking jointly.
The AP loss introduced by 
He \etal~\cite{He2018b} is specially tailored to deal with score ties 
naturally occurring with hamming distances in the context of image hashing. 
Interestingly, the same formulation 
is also proved 
successful for patch matching and retrieval~\cite{He2018}.
Yet their tie-aware formulation poses important convergence problems 
and requires several approximations to be usable in practice. 
In contrast, we propose a straightforward formulation 
of the AP loss that is stable and performs better.
We apply it to image retrieval, which is a rather different task, 
as it involves high-resolution images with significant clutter, 
large viewpoint changes and deeper networks.

Apart from~\cite{He2018b,He2018}, several relaxations or alternative formulations have
been proposed in the literature to allow for the direct optimization of AP
\cite{Behl2015,Hazan2010,He2018,Henderson2016,Mohapatra2014,Song2016,Yue2007}.
Yue \etal \cite{Yue2007} proposed to optimize the AP through a loss-augmented
inference problem \cite{Hazan2010} under a structured learning framework
using linear SVMs. Song \etal \cite{Song2016} then expanded this
framework to work with non-linear models. However, both works assume
that the inference problem in the loss-augmented inference in the
structured SVM formulation can be solved efficiently~\cite{Henderson2016}.
Moreover, their technique requires using a dynamic-programming approach
which requires changes to the optimization algorithm itself, complicating
its general use. 
The AP loss had not been implemented for the general case of deep
neural networks trained with arbitrary learning algorithms until very
recently \cite{He2018,Henderson2016}. Henderson and Ferrari \cite{Henderson2016}
directly optimize the AP for object detection, while He \etal~\cite{He2018} optimize for
patch verification, retrieval, and image matching. % connection AP -> stage wise

In the context of image retrieval, additional hurdles must be cleared.
Optimizing directly for mAP indeed poses memory issues, as 
high-resolution images and very deep networks are typically used at training and test time~\cite{Gordo2016,Radenovic2016}.
To address this, smart multi-stage backpropagation methods have
been developed for the case of image triplets~\cite{Gordo2017}, and we show 
that in our setting slightly more elaborate algorithms can be exploited 
for the same goal.

\section{Method\label{sec:method}}

This section introduces the mathematical framework of the AP-based
training loss and the adapted training procedure we adopt 
for the case of high-resolution images.

\subsection{Definitions}

We first introduce mathematical notations. Let $\mathcal{I}$ denote
the space of images and $\mathcal{S}$ denote the unit hypersphere in
$C$-dimensional space,
\ie $\mathcal{S}=\left\{ x\in\mathbb{R}^{C}|\left\Vert x\right\Vert =1\right\}$.
We extract image embeddings using a deep feedforward network
$f_{\Theta}:\mathcal{I}\rightarrow\mathcal{S}$, where $\Theta$ represents 
learnable parameters of the network. We assume that $f_{\Theta}(\cdot)$ is 
equipped with an $\text{L}_{2}$-normalization output module so that the 
embedding $d_i=f_{\Theta}(I_i)$ has unit norm. The similarity between two 
images can then be naturally evaluated in the embedding space using the 
cosine similarity: 
\begin{equation}
sim(I_i,I_j)=d_i^{\top}d_j \in\left[-1,1\right].\label{eq:sim}
\end{equation}

Our goal is to train the parameters $\Theta$ to rank, for each given query image $I_{q}$, its similarity to every image from a database $\{I_{i}\}_{1\leq i \leq N}$ of size $N$.
After computing the embeddings associated with all images by a forward pass of our network, the similarity
$sim(I_{q},I_{i})=S_{i}^{q}$ of each database item to the query
is efficiently measured in the embedding space using Eq. \eqref{eq:sim}, for all $i$ in $\mathcal{N}=\{1,2,\dots,N\}$.
Database images are then sorted according to their similarities in
decreasing order. Let $R:\mathbb{R}^N\times \mathcal{N}\rightarrow \mathcal{N}$ denote the ranking function, where $R(S^q,i)$ is the index of the $i$-th highest value of $S^q$. By extension, $R(S^q)$ denotes the ranked list of indexes for the database. The quality of the ranking $R(S^q)$ can then be
evaluated with respect to the ground-truth image relevance, denoted
by $Y^{q}$ in $\{0,1\}^{N}$, where $Y^q_i$ is $1$ if $I_i$ is relevant to $I_q$ and $0$ otherwise.

Ranking evaluation is performed with one of the information retrieval
(IR) metrics, such as mAP, F-score, and discounted
cumulative gain. In practice (and despite some shortcomings \cite{APstudy2005}), 
AP has become the \emph{de facto} standard metric for IR when the groundtruth labels are binary. 
In contrast to other ranking metrics such as recall or F-score, 
AP does not depend on a threshold, rank position or number of relevant images, 
and is thus simpler to employ and better at generalizing for different queries. 
We can write the $AP$ as a function of $S^q$ and $Y^q$:
\begin{equation}
\text{AP}(S^{q},Y^{q})=\sum_{k=1}^{N}P_{k}(S^{q},Y^{q})\ \Delta r_{k}(S^{q},Y^{q}),\label{eq:ap}
\end{equation}
where $P_k$ is the precision at rank $k$, \ie the proportion of relevant
items in the $k$ first indexes, which is given by:
\begin{equation}
P_{k}(S^{q},Y^{q})=\frac{1}{k}\sum_{i=1}^{k}\sum_{j=1}^N Y_{j}^{q}\indic [R(S^{q},i)=j] \label{eq:prec},
\end{equation}
$\Delta r_k$ is the incremental recall from ranks $k-1$ to $k$, \ie the proportion of the total $N^q=\sum_{i=1}^N Y^q_i$ relevant items found at rank $k$, which is given by:
\begin{equation}
\Delta r_{k}(S^{q},Y^{q})=\frac{1}{N^q}\sum_{j=1}^N Y_{j}^{q}\indic [R(S^{q},k)=j],\label{eq:delta_rec}
\end{equation}
and $\indic[\cdot]$ is the indicator function.

\subsection{Learning with average precision}
Ideally, the parameters of $f_{\Theta}$ should be trained using stochastic
optimization such that they maximize AP on the training set. This is not feasible for the original AP formulation, because of the presence of the indicator function $\indic[\cdot]$. 
Specifically, the function $R \mapsto \indic[R=j]$ has derivative w.r.t. $R$ equal to zero for all $R \ne 0$ and its derivative is undefined at $R=0$. This derivative thus provides no information for optimization.

Inspired by listwise losses developed for histograms
\cite{Ustinova2016}, an alternative way of computing the AP has recently
been proposed and applied for the task of descriptor hashing \cite{He2018b}
and patch matching \cite{He2018}. 
The key is to train with a relaxation of the AP, obtained by replacing
the hard assignment $\indic$ by a function $\delta$, whose derivative
can be backpropagated, that soft-assigns similarity values into a fixed
number of bins.
Throughout this section, for simplicity, we will refer to functions
differentiable almost everywhere as differentiable.

\vspace{-2mm}
\paragraph{Quantization function.}
For a given positive integer $M$, we partition the interval $\left[-1,1\right]$ 
into $M-1$ equal-sized intervals, each of measure $\Delta=\frac{2}{M-1}$ and 
limited (from right to left) by \emph{bin centers} $\{b_m\}_{1\leq m \leq M}$,
where ${b_m=1-(m-1)\Delta}$. 
In Eq. \eqref{eq:ap} we calculate precision and incremental recall at every 
rank $k$ in $\{1,\dots,N\}$. The first step of our relaxation is to, instead,
compute these values at each bin: 
\begingroup
\vspace{-2mm}
\addtolength{\jot}{0.5em}
\begin{align}
P^{bin}_m (S^q,Y^q)&= \dfrac{\sum_{m'=1}^m\sum_{i=1}^N Y^q_i \indic [S^q_i\in \bar{b}_{m'}]} {\sum_{m'=1}^m\sum_{i=1}^N \indic [S^q_i\in \bar{b}_{m'}]}, \\
\Delta r^{bin}_m (S^q,Y^q)&= \dfrac{\sum_{i=1}^N Y^q_i \indic [S^q_i\in \bar{b}_m]}{N^q},
\end{align}
\endgroup
where the interval $\bar{b}_m=[\text{max}(b_m-\Delta,-1),\text{min}(b_m+\Delta,1))$ denotes the $m$-th bin.

The second step is to use a soft assignment as replacement of the indicator.
Similarly to \cite{He2018b}, we define the function 
$\delta:\mathbb{R}\times \{1,2,\dots,M\}\rightarrow [0,1]$ such that each
$\delta(\cdot,m)$ is a triangular kernel centered around $b_m$ and width
$2\Delta$, \ie
\begin{equation}
\delta(x,m) = \max \left( 1-\frac{|x-b_m|}{\Delta}, 0\right).\label{eq:binning}
\end{equation}
$\delta(x,m)$ is a soft binning of $x$ that approaches the indicator function $\indic[x\in \bar{b}_m]$ when $M\rightarrow \infty$ while being differentiable w.r.t. $x$:
\begin{equation}
\frac{\partial \delta(x,m)}{\partial x}=-\dfrac{\text{sign}(x-b_m)}{\Delta}\indic \left[|x-b_m|\leq \Delta\right].
\end{equation}
By expanding the notation, $\delta(S^q,m)$ is a vector in $[0,1]^N$ that indicates the soft assignment of $S^q$ to the bin $\bar{b}_m$.

Hence, the \emph{quantization} $\{\delta(S^q,m)\}_{i=1}^{N}$ of $S^q$ is 
a smooth replacement of the indicator function. This allows us to 
recompute approximations of precision and incremental recall as 
function of the quantization, as presented previously in Eq.~\eqref{eq:prec}
and Eq.~\eqref{eq:delta_rec}. Thus, for each bin $m$, the quantized 
precision $\hat{P}_m$ and incremental recall $\Delta \hat{r}_m$ are 
computed as:
\begingroup
\vspace{-2mm}
\addtolength{\jot}{0.5em}
\begin{align}
\hat{P}_m (S^q,Y^q)&= \dfrac{\sum_{m'=1}^m\delta(S^q,m')^\top Y^q }{\sum_{m'=1}^m \delta(S^q,m')^\top \text{\textbf{1}}}, \\
\Delta \hat{r}_m (S^q,Y^q)&= \dfrac{\delta(S^q,m)^\top Y^q}{N^q},
\end{align}
\endgroup
and the resulting \emph{quantized average precision}, denoted by $\text{AP}_Q$, is a smooth function w.r.t. $S^q$, given by:
\begin{equation}
\AP(S^q, Y^q) = \sum_{m=1}^{M}\hat{P}_{m}(S^{q},Y^{q})\ \Delta \hat{r}_{m}(S^{q},Y^{q}).\label{eq:apquant}
\end{equation}

\begin{figure*}[t!]
    \centering
%	\vspace*{-3mm}
	% trim={<left> <lower> <right> <upper>}
	\includegraphics[trim={0 0 0 6mm},width=0.95\linewidth]{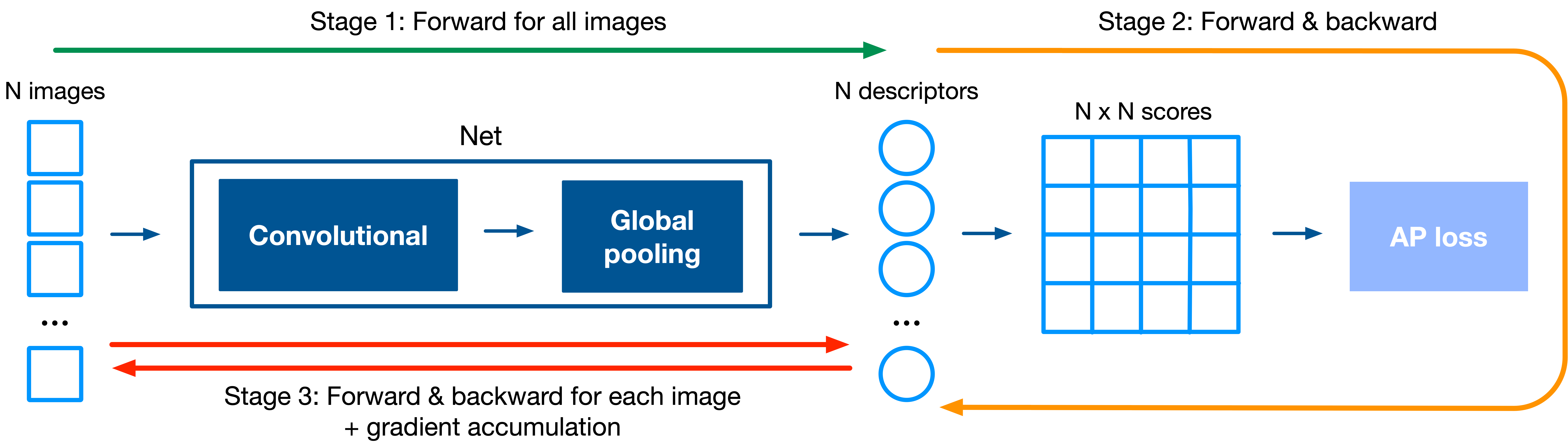}
	\caption{
        %\scriptsize
        Illustration of the multistaged network optimization. During
		the \textbf{\color{OliveGreen}first stage}, we compute the descriptors
		of all batch images, discarding the intermediary tensors in the memory.
		In the \textbf{\color{Orange}second stage}, we compute the score matrix
		$S$ (Eq. \ref{eq:sim}) and the $\mAP$ loss $\ell=L(D,Y)$, and we compute
		the gradient of the loss w.r.t. the descriptors. During the
		\textbf{\color{Red}last stage}, given an image in the batch, we recompute
		its descriptor, this time storing the intermediate tensors, and use the
		computed gradient for this descriptor to continue the backpropagation
		through the network. Gradients are accumulated, one image at a time,
		before finally updating the network weights.}
	\label{fig:multis}
%	\vspace*{-4mm}
\end{figure*}

\vspace{-3mm}
\paragraph{Training procedure.}
The training procedure and loss are defined as follows. Let $\mathcal{B}=\{I_{1},\ldots,I_{B}\}$
denote a batch of images with labels $[y_{1},\ldots, y_{B}]\in\mathbb{N}^{B}$,
and $D=[d_{1},\ldots,d_{B}]\in\mathcal{S}^{B}$ their corresponding
descriptors. During each training iteration, we compute the 
mean $\text{AP}_{Q}$ over the batch. To that goal we consider each of the batch
images as a potential query and compare it to all other batch images.
The similarity scores for query $I_{i}$ are denoted by $S_{i}\in[-1,1]^{B}$,
where $S_{ij}=d_{i}^{\top}d_{j}$ is the similarity with image $I_{j}$.
Meanwhile, let $Y_{i}$ denote the associated binary ground-truth,
with $Y_{ij}=\indic[y_{i}=y_{j}]$. We compute the quantized mAP, denoted by
$\mAP$, for this batch as:
\begin{equation}
\mAP(D,Y)=\frac{1}{B}\sum_{i=1}^{B}\AP\left(d_{i}^{\top}D,Y_{i}\right)\label{eq:mAP}
\end{equation}
Since we want to maximize the mAP on the training set, the loss is
naturally defined as $L(D,Y)=1-\mAP(D,Y)$.

\subsection{Training for high-resolution images}
He \etal \cite{He2018} have shown that, in the context of patch
retrieval, top performance is reached for large batch sizes. In the
context of image retrieval, the same approach cannot be applied directly.
Indeed, the memory occupied by a batch is several orders
of magnitude larger then that occupied by a patch, making the backpropagation intractable
on any number of GPUs. This is because ($\romannumeral 1$) high-resolution images
are typically used to train the network, and ($\romannumeral 2$) the network used
in practice is much larger (ResNet-101 has around 44M parameters, whereas % $\sim$
the L2-Net used in \cite{He2018} has around 26K).
Training with high-resolution images is known
to be crucial to achieve good performance, both at training and test
time \cite{Gordo2017}. Training images fed to the network typically have a
resolution of about 1Mpix (compared to $51 \times 51$ patches in \cite{He2018}). 

By exploiting the chain rule, we design a \textit{multistaged backpropagation} %an algorithm
that solves this memory issue and allows the training of a network of arbitrary
depth, and with arbitrary image resolution and batch size without approximating the loss. The algorithm is illustrated in Fig.~\ref{fig:multis}, and consists of three stages.

During the first stage, we compute the descriptors of all batch images,
discarding the intermediary tensors in the memory (\ie
in evaluation mode). 
In the second stage, we compute the score matrix $S$ (Eq. \ref{eq:sim})
and the loss $\ell=L(D,Y)$, and we compute the gradient
of the loss w.r.t. the descriptors $\frac{\partial\ell}{\partial d_{i}}$.
In other words, we stop the backpropagation before entering the network.
Since all tensors considered are compact (descriptors, score matrix),
this operation consumes little memory. 
During the last stage, we recompute the image descriptors, this time storing the intermediary tensors.
Since this operation occupies a lot of memory, we perform this operation
image by image. Given the descriptor $d_{i}$ for the image $I_{i}$
and the gradient for this descriptor $\frac{\partial\ell}{\partial d_{i}}$,
we can continue the backpropagation through the network. We thus accumulate
gradients, one image at a time, before finally updating the network
weights. 
Pseudo-code for multistaged backpropagation can be found in the supplementary material.

\section{Experimental results\label{sec:xp}}

We first discuss the different datasets used in our experiments. 
We then report experimental results on these datasets, studying 
key parameters of the proposed method and comparing with the state of the art.

\subsection{Datasets}\label{ss:datasets}

\paragraph{Landmarks.}
The original Landmarks dataset \cite{Babenko2014} contains 213,678
images divided into 672 classes. However, since this dataset has been
created semi-automatically by querying a search engine, it contains
a large number of mislabeled images. 
In \cite{Gordo2016}, Gordo \etal proposed an automatic 
cleaning process to clean this dataset to use with their retrieval
model, and made the cleaned dataset public. This \textit{Landmarks-clean} 
dataset contains 42,410 images and 586 landmarks, and it is the 
version we use to train our model in all our experiments.

\begin{figure*}[t!]
	\centering
	\includegraphics[trim={0 -0mm 5mm 0},clip,width=0.95\linewidth]{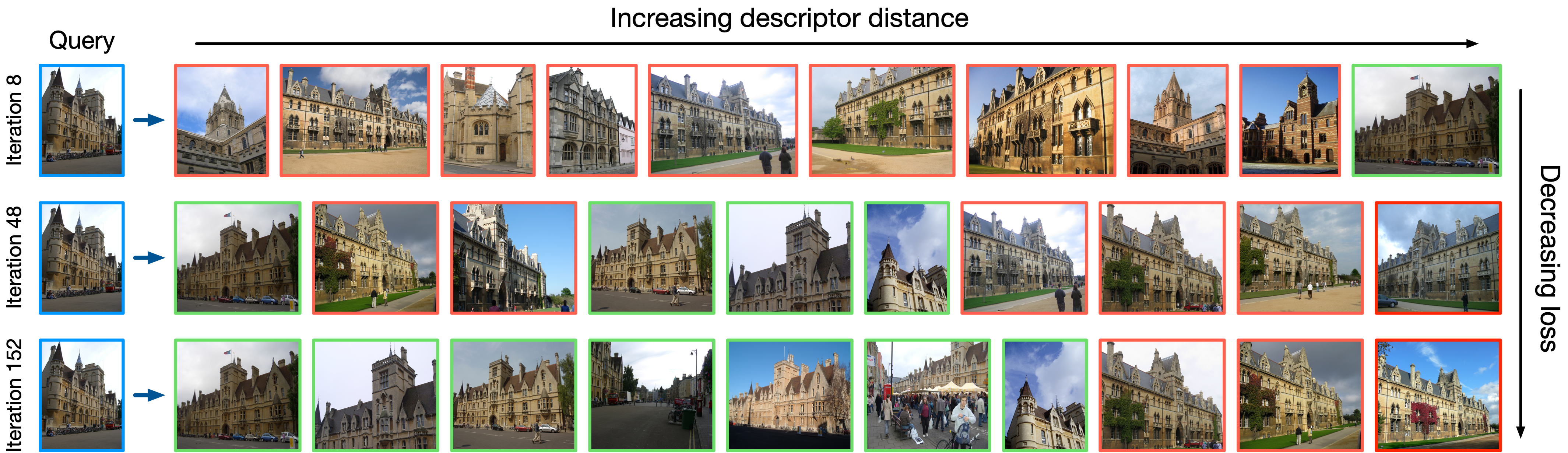}
	\caption{Example queries for \ROxford~and illustration for the evolution of the listwise loss during training. As 
		training progresses, images get sorted by descriptor distance to the query image,
		decreasing the value for the loss. }
	\label{fig:ranking}
	\vspace*{-4mm}
\end{figure*}

\vspace{-3mm}
\paragraph{Oxford and Paris Revisited.}
Radenovi\'{c} {\etal} have recently revised the Oxford~\cite{Philbin07}
and Paris~\cite{Philbin08} buildings datasets correcting annotation 
errors, increasing their sizes, and providing new protocols for their 
evaluation \cite{Raden2018}.
The Revisited Oxford (\ROxford) and Revisited Paris (\RParis) datasets
contain 4,993 and 6,322 images respectively, with 70 additional
images for each that are used as queries (see Fig.~\ref{fig:ranking} for example queries).
These images are further labeled according to the difficulty in 
identifying which landmark they depict. 
These labels are then used to determine three evaluation
protocols for those datasets: \textit{Easy},
\textit{Medium}, and \textit{Hard}.
Optionally, a set of 1 million distractor images (\ROM) can be added to each dataset
to make the task more realistic.
Since these new datasets are essentially updated versions of the
original Oxford and Paris datasets, with the same characteristics but 
more reliable ground-truth, we use these revisited versions in our 
experiments.

\subsection{Implementation details and parameter study}

We train our network using stochastic gradient with Adam~\cite{Kingma2014}
on the public Landmarks-clean dataset of \cite{Gordo2016}.
In all experiments we use ResNet-101~\cite{He2016} pre-trained on 
ImageNet~\cite{Russakovsky2015} as a backbone. We append a generalized-mean
pooling (GeM) layer~\cite{Radenovic2018a} which was recently 
shown to be more effective than R-MAC pooling~\cite{Gordo2017,Tolias16}.
The GeM power is trained using backpropagation along the other weights.
Unless stated otherwise, we use the following parameters:
we set weight decay to $10^{-6}$, and apply standard data augmentation 
(\eg color jittering, random scaling, rotation and cropping).
Training images are cropped to a fixed size of $800 \times 800$,
but during test, we feed the original images (unscaled and undistorted) to the network. 
We tried using multiple scales at test time, but did not observe any significant 
improvement. Since we operate at a single scale, this essentially makes our descriptor 
extraction about 3 times faster than state-of-the-art methods~\cite{Gordo2017,Raden2018}
for a comparable network backbone.
We now discuss the choice of other parameters based on different experimental studies.

\paragraph{Learning rate.} 
We find that the highest learning rate that does not result in divergence gives best results.
We use a linearly decaying rate starting from $10^{-4}$ that decreases to $0$ after 200 iterations.

\begin{figure}
	\centering
	\includegraphics[width=0.82\linewidth]{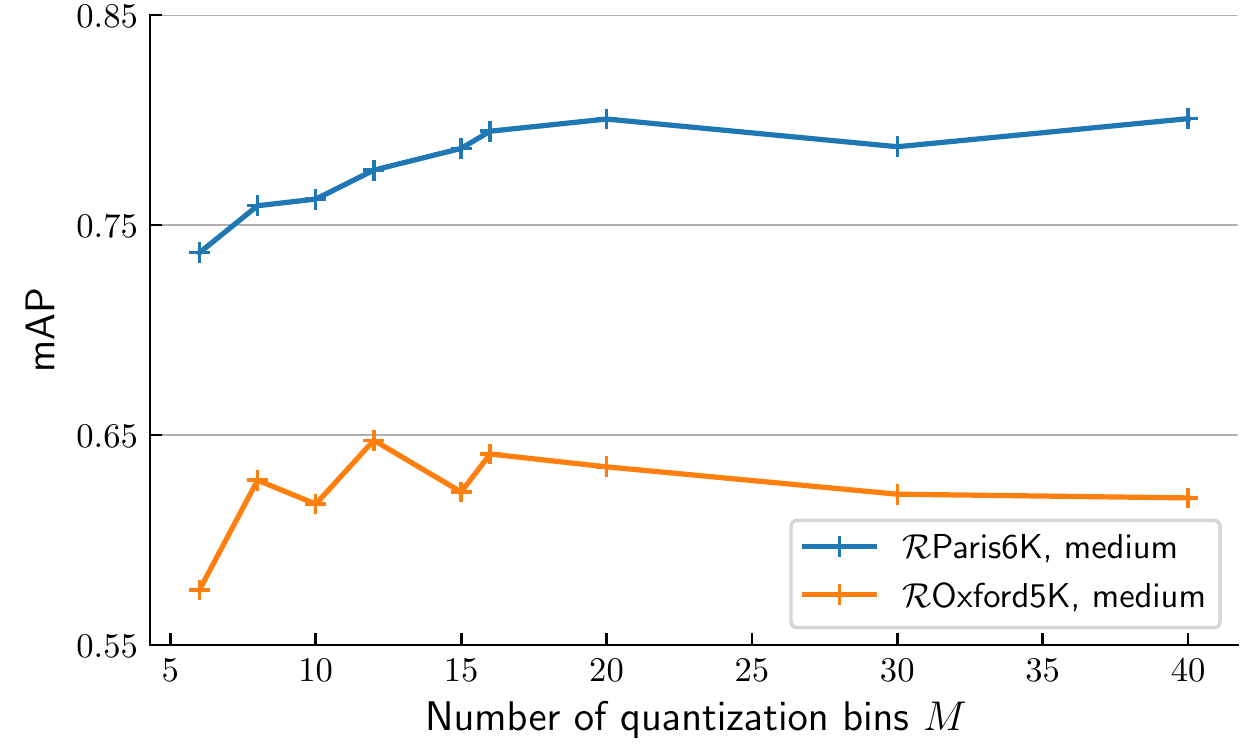}
	\caption{\label{fig:quantbin}
		mAP for the \textit{medium} benchmarks of \RParis~and \ROxford~for different 
		numbers	of quantization bins $M$ (Eq.~\ref{eq:binning}), 
		showing this parameter has little impact on performance.}
	\vspace{\floatsep}
	\includegraphics[width=0.82\linewidth]{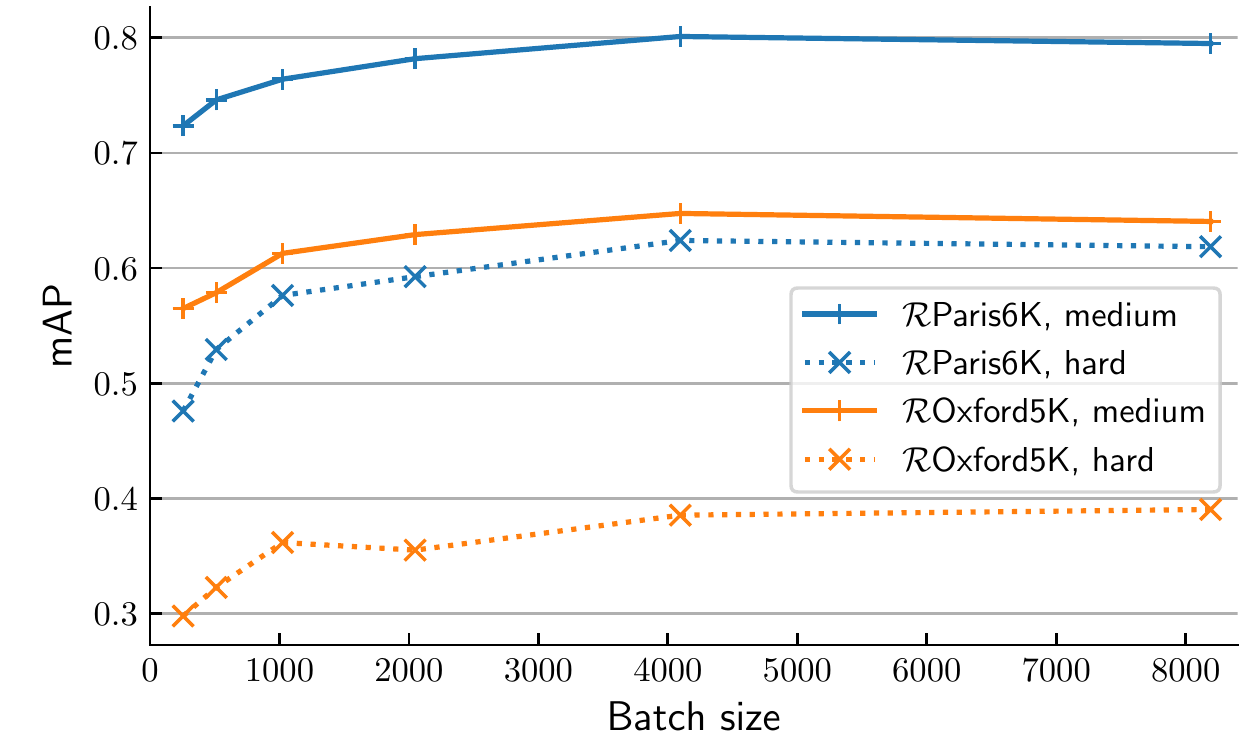}
	\caption{\label{fig:batchsize}
		mAP for different batch sizes $B$. Best results 
		are obtained with large batch sizes. 
			}
	\vspace{\floatsep}
	\includegraphics[width=0.92\linewidth]{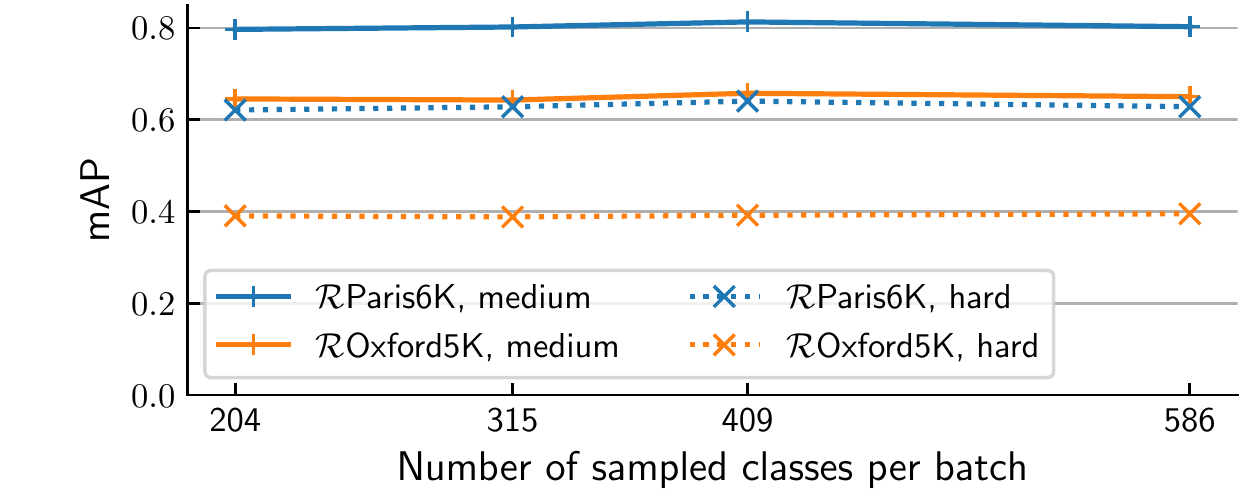}
	\caption{\label{fig:nc}
		Assuming that it could be beneficial for the model to see images from different classes
		at each iteration, we construct each batch by sampling images from a limited random set of classes.
		This has little to no effect on the final performance. 
	}
	\vspace{-2mm}
\end{figure}

\vspace{-4mm}
\paragraph{Batch size.} 
As pointed by~\cite{He2018}, we find that larger batch sizes lead to better results (see Fig.~\ref{fig:batchsize}). 
The performance saturates beyond 4096, and training significantly slows down as well. 
We use $B=4096$ in all subsequent experiments.

\begin{figure}[t!]
	\vspace{-1mm}
	\centering
	\includegraphics[width=0.84\linewidth]{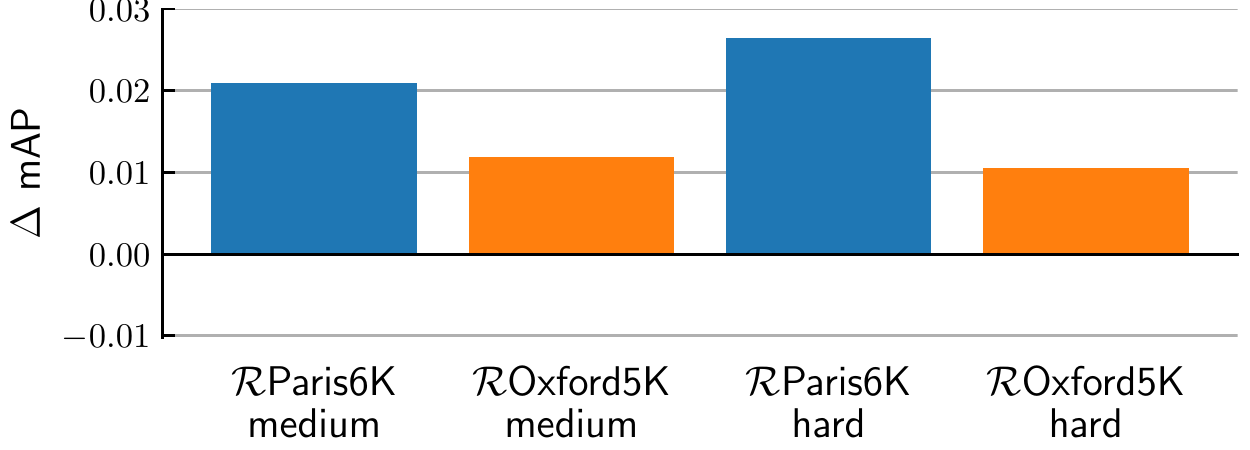}
	\caption{\label{fig:clsbalance}
		Balancing AP weights when computing the AP loss, such that all classes are weighted
		equally within a batch, brings a performance improvement of 1 to 3\%. 
	}
\end{figure}

\vspace{-4mm}
\paragraph{Class sampling.}
We construct each batch by sampling random images from each dataset class 
(all classes are hence represented in a single batch). 
We also tried sampling classes but did not observe any difference (see Fig.~\ref{fig:nc}). 
Due to the dataset imbalance, certain classes are constantly over-represented at the batch level. 
To counter-balance this situation, we introduce a 
weight in Eq.~\eqref{eq:mAP} to weight equally all classes inside a batch. 
We train two models with and without this option and present the results in Fig.~\ref{fig:clsbalance}.
The improvement in mAP with class weighting is around +2\% and shows the importance of this balancing.

\begin{figure}[t!]
    \centering
    \vspace{-3mm}
    %trim={<left> <lower> <right> <upper>}
    \includegraphics[trim={0 0 0 4mm}, clip,  width=0.84\linewidth]{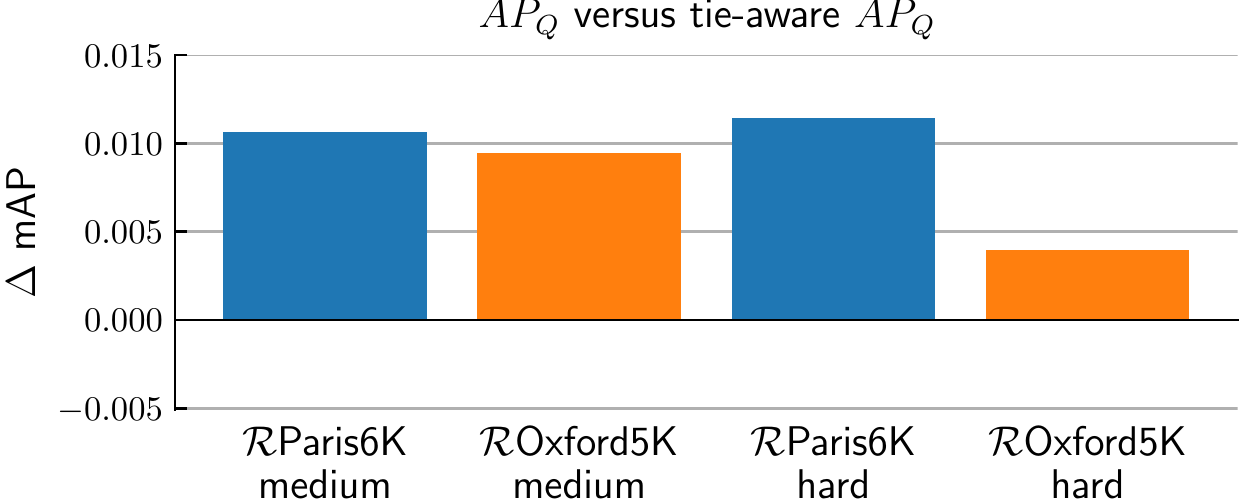}
    \vspace{1mm}
    \caption{\label{fig:tap}
        Improvement in mAP from mean $\text{mAP}_{\text{T}}$ loss to mean $\mAP$ loss.
        Our AP formulation results in a small but constant improvement
        with respect to the tie-aware AP formulation from~\cite{He2018b}.
    }
    \vspace{-4.25mm}
\end{figure}

\vspace{-4mm}
\paragraph{Tie-aware AP.} 
In~\cite{He2018b}, a tie-aware version of the $\mAP$ loss
is developed for the specific case of ranking integer-valued Hamming distances.
\cite{He2018} uses the same version of AP for real-valued Euclidean distances. We trained
models using simplified tie-aware AP loss (see Appendix F.1 of \cite{He2018b}), denoted by $\text{mAP}_{\text{T}}$ in addition to the original $\mAP$ loss. We write the $\text{mAP}_{\text{T}}$ similarly to Eq. \eqref{eq:apquant}, but replacing the precision by a more accurate approximation:
\vspace{-1mm}
\begin{equation}
\small
\hat{P}_m (S^q,Y^q) = \dfrac{1+ \delta(S^q,m)^\top Y^q +2\sum_{m'=1}^{m-1} \delta(S^q,m')^\top Y^q }{1+\delta(S^q_i,m)^T \text{\textbf{1}}+2\sum_{m'=1}^{m-1}\delta(S^q_i,m')^\top\text{\textbf{1}}}
\end{equation}
The absolute difference in mAP is presented in Fig.~\ref{fig:tap}. We find that
$\mAP$ loss, straightforwardly derived from the definition of AP,
consistently outperforms the tie-aware formulation by a small but significant margin.
This may be due to the fact that the tie-aware formulation used in practical
implementations is in fact an approximation of the theoretical tie-aware AP
(see appendix in \cite{He2018b}).
We use the $\mAP$ loss in all subsequent experiments.

\clearpage

\def\dagg{\textsuperscript{\textdagger}}
\def\daggd{\textsuperscript{\textdaggerdbl}}
\newcommand{\mr}[2]{\multirow{#1}{*}[-1mm]{\makecell{#2}}}

\begin{table*}[b!]
    \vspace*{-2mm}
	\caption{Ceteris paribus analysis on the loss. 
}
	\vspace*{1mm}
	\label{tab:efficiency}
	\small
	\centering
	\setlength{\tabcolsep}{2.5pt}
	\begin{threeparttable}
		\begin{tabularx}{\textwidth}{lcccccccccccc}
\toprule
		    & \multicolumn{2}{c}{\textbf{Medium}} && \multicolumn{2}{c}{\textbf{Hard}}&&\mr{2}{Number of\\Backwards}
                                                                                       &\mr{2}{Number of\\Forwards}
                                                                                       &\mr{2}{Number of\\Updates}
                                                                                       &\mr{2}{Number of\\hyper-param.\daggd} 
                                                                                       &\mr{2}{Extra lines\\of code} 
                                                                                       &\mr{2}{Training\\time}      \\ 
\cmidrule[0.5pt]{2-3} \cmidrule[0.5pt]{5-6}
Method      & \textbf{$\RR$Oxf} & \textbf{$\RR$Par} &  & \textbf{$\RR$Oxf} & \textbf{$\RR$Par} & \phantom{a}  &     & & & &     \\ 
\midrule
GeM (AP) [ours]                     & \bf{67.4}  & \bf{80.4}  && \bf{42.8}  & \bf{61.0}   && \bf{\tb{819K}}  & \bf{\tb{1638K}}  & \bf{\tg{200}}  & \bf{\tr{2}}  & \bf{15}  & \bf{1} day \\
GeM (TL-64) [ours]                  & 64.9  & 78.4  && 41.7  & 58.7   && 1572K & 2213K  &  8192  &  6 & 175 (HNM) &  3 days \\ 
GeM (TL-512) [ours]                 & 65.8  & 77.6  && 41.3  & 57.1   && 2359K & 3319K  &  1536  &  6 & 175 (HNM) &  3 days \\ 
GeM (TL-1024) [ours]                & 65.5  & 78.6  && 41.1  & 59.1   && 3146K & 4426K  &  1024  &  6 & 175 (HNM) &  3 days \\ 
\midrule
R-MAC (TL)\dagg~\cite{Gordo2017}    & 60.9  & 78.9  && 32.4  & 59.4   && \tb{1536K} & \tb{3185K} & \tg{8000} & \tr{6}  & 100+ (HNM) & 4 days \\
GeM (CL)\dagg~\cite{Radenovic2018a} & 64.7  & 77.2  && 38.5  & 56.3   && \tb{1260K} & \tb{3240K} & \tg{36000}& \tr{7} & 46 (HNM) & 2.5 days  \\
\bottomrule
		\end{tabularx}
		\begin{tablenotes}[flushleft]
			\footnotesize
	    \item \dagg For the sake of completeness, we include metrics from \cite{Gordo2017} and \cite{Radenovic2018a} in the last two rows of the table even though they are not exactly comparable due the usage of different training sets or whitening and pooling mechanisms. 
	    \daggd See supplementary material for a listing of those parameters.
		\end{tablenotes}
\end{threeparttable}
\end{table*}

\paragraph{Score quantization.} 
Our $\mAP$ loss depends on the number of quantization
bins $M$ in Eq.~\eqref{eq:binning}. We plot the performance achieved for different
values of $M$ in Fig.~\ref{fig:quantbin}. In agreement to previous findings~\cite{He2018,Ustinova2016}, 
this parameter has little impact on the performance. We use $M=20$ quantization bins in all other experiments.

\paragraph{Descriptor whitening.}
As it is common practice \cite{Jegou2012,Raden2018}, we whiten
our descriptors before evaluation. First, we learn a PCA from
descriptors extracted from the Landmarks dataset. Then, we use
it to normalize the descriptors for each benchmark dataset. As
in~\cite{Jegou2012}, we use a square-rooted 
PCA. We use whitening in all subsequent experiments.

\subsection{Ceteris paribus analysis}
In this section, we study in more details the benefits of using the proposed listwise loss
with respect to a state-of-the-art loss.
For this purpose, we replace in our approach the proposed $\mAP$ loss by the triplet loss (TL)
accompanied by hard negative mining (HNM) as described in \cite{Gordo2017} 
(\ie using batches of 64 triplets).
We then re-train the model, keeping the pipeline identical and
separately re-tuning all hyper-parameters, such as the learning rate and the weight decay. 
Performance after convergence is presented in the first two lines of Table~\ref{tab:efficiency}. Our implementation of the 
triplet loss, denoted as ``GeM (TL-64)'', is on par or better than~\cite{Gordo2017}, 
which is likely due to switching from R-MAC to GeM pooling~\cite{Radenovic2018a}.
More importantly, we observe a significant improvement when using the proposed $\mAP$ loss
(up to 3\% mAP) even though no hard-negative mining scheme is employed.
We stress that training with the triplet loss using larger batches 
(\ie seeing more triplets before updating the model), denoted as GeM (TL-512) and GeM (TL-1024),
does not lead to increased performance as shown in Table~\ref{tab:efficiency}. 
Note that TL-1024 corresponds to seeing $1024\times 3=3072$ images per model update,
which roughly corresponds to a batch size $B=4096$ with our $m\AP$ loss.
This shows that the good performance of the $\mAP$ loss is not solely 
due to using larger training batches. 

We also indicate in Table~\ref{tab:efficiency} the training effort for each method 
(number of backward and forward passes, number of updates, total training time) 
as well as the number of hyper-parameters and extra lines of code they require with respect to a base implementation 
(see supplementary material for more details).
As it can be observed, our approach leads to considerably fewer
weight updates than when using a local loss. It also leads
to a significant reduction in forward and backward passes. 
This supports our claim that considering all images at once with a listwise loss is much more effective.
Overall, our model is trained 3 times faster than with a triplet loss.

The methods can also be compared regarding the amount of engineering involved. 
For instance, hard negative mining can easily require hundreds of additional 
lines, and comes with many \tr{extra hyper-parameters}.
In contrast, the PyTorch code for the proposed backprop is only 5 lines longer
than a normal backprop, the AP loss itself can be implemented in 10 lines,
and our approach requires only 2 hyper-parameters
(the number of bins $M$ and batch size $B$),
that have safe defaults and are not very sensitive to changes.

\subsection{Comparison with the state of the art}

\begin{table*}[t]
    \vspace*{-1mm}
	\caption{Performance evaluation (mean average precision) for \ROxford~and \RParis. }
	\vspace*{-3mm}
	\label{tab:sota}
	\begin{center}
		\setlength{\tabcolsep}{.45em}
		\begin{threeparttable}
		\begin{tabular}{@{}lccccccccc@{}}
			\toprule
			& \multicolumn{4}{c}{\textbf{Medium}} && \multicolumn{4}{c}{\textbf{Hard}} \\
               \cmidrule[0.5pt]{2-5} \cmidrule[0.5pt]{7-10}

			& \textbf{$\RR$Oxf} & \textbf{$\RR$Oxf+1M} & \textbf{$\RR$Par} & \textbf{$\RR$Par+1M} & \phantom{ab} & \textbf{$\RR$Oxf} & \textbf{$\RR$Oxf+1M} & \textbf{$\RR$Par} & \textbf{$\RR$Par+1M}\\
			\midrule[0.6pt]
\textbf{Local descriptors} \\
			HesAff-rSIFT-ASMK$^{*}$ + SP \cite{Raden2018}& 60.6&	46.8& 61.4&	42.3&& 36.7&	26.9& 35.0&	16.8\\
			DELF-ASMK$^{*}$ + SP \cite{Landmarks2017}& \textbf{67.8}&	\textbf{53.8}& \textbf{76.9}&	\textbf{57.3}&& \textbf{43.1}&	\textbf{31.2}& \textbf{55.4}&	\textbf{26.4}\\ \midrule[0.2pt]
\textbf{Global representations} \\
			 MAC (O) \cite{Tolias16} & 41.7 & 24.2 & 66.2 & 40.8 && 18.0&5.7& 44.1 & 18.2\\
			SPoC (O) \cite{Babenko2015} & 39.8 & 21.5 & 69.2 & 41.6 && 12.4&2.8& 44.7 & 15.3\\
			CroW (O) \cite{Kalantidis2016} & 42.4 & 21.2 & 70.4 & 42.7 && 13.3 & 3.3 & 47.2 & 16.3\\
			R-MAC (O) \cite{Tolias16} & 49.8 & 29.2 & 74.0 & 49.3 &&	18.5& 4.5& 52.1&21.3\\
			R-MAC (TL) \cite{Gordo2017} & 60.9 & 39.3 & 78.9 & \textbf{54.8} && 32.4&12.5&59.4&	\textbf{28.0}\\
			GeM (O) \cite{Radenovic2018a} & 45.0 & 25.6 & 70.7 & 46.2 && 17.7&	4.7 & 48.7 & 20.3\\
			GeM (CL) \cite{Radenovic2018a} & 64.7 & 45.2 & 77.2 & 52.3 && 38.5& 19.9 & 56.3 & 24.7\\
			GeM (AP) [ours] & \textbf{67.5} & \textbf{47.5} & \textbf{80.1} & 52.5 && \textbf{42.8} & \textbf{23.2} & \textbf{60.5} & 25.1 \\

			\midrule[0.2pt]
\textbf{Query expansion} \\
            R-MAC (TL) + $\alpha$QE \cite{Gordo2017} &	64.8&45.7&82.7&\textbf{61.0}&&36.8&19.5&65.7&\textbf{35.0}\\
			GeM (CL) + $\alpha$QE \cite{Radenovic2018a} &67.2&49.0&	80.7&58.0&&40.8&	24.2&61.8&31.0\\
			GeM (AP) + $\alpha$QE [ours] & \textbf{71.4} & \textbf{53.1} & \textbf{84.0} & 60.3 && \textbf{45.9} & \textbf{26.2} & \textbf{67.3} & 32.3 \\
			\midrule[0.2pt]
            \bottomrule
		\end{tabular}
		\begin{tablenotes}[flushleft]
			\item \footnotesize All global representations are learned from a ResNet-101 backbone, with varying pooling layers and fine-tuning losses. Abbreviations: 
			(O) off-the-shelf features;
			(CL) fine-tuned with contrastive loss; 
			(TL) fine-tuned with triplet loss; 
			(AP) fine-tuned with mAP loss (ours);
			(SP) spatial verification with RANSAC;
			($\alpha$QE) weighted query expansion \cite{Radenovic2018a}.
		\end{tablenotes}
		\end{threeparttable}
	\end{center}
	\vspace*{-8mm}
\end{table*}

We now compare the results obtained by our model 
with the state of the art.
The top part of Table~\ref{tab:sota} summarizes the performance of the best-performing methods 
on the datasets listed in section \ref{ss:datasets} without query 
expansion. 
We use the notation of~\cite{Raden2018} as this helps us to clarify important aspects about each method.
Namely, generalized-mean pooling~\cite{Radenovic2018a} is denoted by GeM and the R-MAC pooling~\cite{Tolias16} is denoted by R-MAC.
The type of loss function used to train the model is denoted by (CL) for the contrastive loss,
(TL) for the triplet loss, (AP) for our $\mAP$ loss, and (O) if no loss is used (\ie off-the-shelf features). 

Overall, our model 
outperforms the state of the art by 1\% to 5\% on most of the datasets and protocols. 
For instance, our model is more than 4 points ahead 
of the best reported results~\cite{Radenovic2018a} on the hard protocol of \ROxford~and \RParis. 
This is remarkable since our model uses a single scale at test time (\ie the original test images), 
whereas other methods boost their performance by pooling image descriptors computed at several scales.
In addition, our network does not undergo any special pre-training step
(we initialize our networks with ImageNet-trained weights), 
again unlike most competitors from the state of the art. 
Empirically we observe that the $\mAP$ loss renders such pre-training stages obsolete.
Finally, the training time is also considerably reduced: training our model from scratch takes
a few hours on a single P40 GPU.
Our models will be made publicly available to download upon acceptance.

We also report results with query expansion (QE)
at the bottom of Table~\ref{tab:sota}, as
is common practice in the literature~\cite{Arandjelovic2012,Gordo2017,Raden2018,Radenovic2018a}.
We use the $\alpha$-weighted versions~\cite{Radenovic2018a} with
$\alpha=2$ and $k=10$ nearest neighbors.
Our model with QE outperforms other methods also using QE in 6 protocols out of 8.
We note that our results are on par with those of the method proposed by Noh \etal~\cite{Landmarks2017}, which is based on local descriptors.
Even though our method relies on global descriptors, hence lacking any geometric verification, 
it still outperforms it on the \ROxford~and \RParis~datasets without added distractors.

\section{Conclusion\label{sec:Conclusion}}

In this paper, we proposed the application of a listwise ranking 
loss to the task of image retrieval. We do so by directly optimizing
a differentiable relaxation of the mAP, that we call $\mAP$. In contrast to
the standard loss functions used for this task, $\mAP$
does not require expensive mining of image samples or careful pre-training.
Moreover, we train our models efficiently using a multistaged optimization scheme
that allows us to learn models on high-resolution images with arbitrary
batch sizes, and achieve state-of-the-art results in multiple benchmarks.
We believe our findings can guide the development of better models
for image retrieval by showing the benefits of optimizing for the
target metric.
Our work also encourages the exploration of image retrieval beyond 
the instance level, by leveraging a metric that can learn from 
a ranked list of arbitrary size at the same time, instead of
relying on local rankings.

\section*{Acknowledgments}
We thank Dr. Christopher Dance who provided insight and expertise that greatly assisted the research and improved the manuscript.

{\small{}{}{}{}  \bibliographystyle{ieee}
\bibliography{egbib}

\begin{thebibliography}{10}\itemsep=-1pt

\bibitem{Arandjelovic15}
R.~Arandjelovi\'c, P.~Gronat, A.~Torii, T.~Pajdla, and J.~Sivic.
\newblock {NetVLAD}: {CNN} architecture for weakly supervised place
  recognition.
\newblock {\em CoRR}, abs/1511.07247, 2015.

\bibitem{Arandjelovic2012}
R.~Arandjelovic and A.~Zisserman.
\newblock Three things everyone should know to improve object retrieval.
\newblock In {\em CVPR}, 2012.

\bibitem{Babenko2015}
A.~Babenko and V.~Lempitsky.
\newblock Aggregating local deep features for image retrieval.
\newblock In {\em ICCV}, 2015.

\bibitem{Babenko2014}
A.~Babenko, A.~Slesarev, A.~Chigorin, and V.~Lempitsky.
\newblock Neural codes for image retrieval.
\newblock In {\em ECCV}, 2014.

\bibitem{Behl2015}
A.~Behl, P.~Mohapatra, C.~V. Jawahar, and M.~P. Kumar.
\newblock Optimizing average precision using weakly supervised data.
\newblock {\em IEEE TPAMI}, 2015.

\bibitem{Burges2005}
C.~Burges, T.~Shaked, E.~Renshaw, A.~Lazier, M.~Deeds, N.~Hamilton, and
  G.~Hullender.
\newblock Learning to rank using gradient descent.
\newblock In {\em ICML}, 2005.

\bibitem{Cakir2017}
F.~Cakir, K.~He, S.~A. Bargal, and S.~Sclaroff.
\newblock {MIHash}: Online hashing with mutual information.
\newblock In {\em ICCV}, 2017.

\bibitem{Cao2007}
Z.~Cao, T.~Qin, T.-Y. Liu, M.-F. Tsai, and H.~Li.
\newblock Learning to rank: from pairwise approach to listwise approach.
\newblock In {\em ICML}, 2007.

\bibitem{Chen2018}
H.~Chen, Y.~Wang, Y.~Shi, K.~Yan, M.~Geng, Y.~Tian, and T.~Xiang.
\newblock Deep transfer learning for person re-identification.
\newblock In {\em International Conference on Multimedia Big Data}, 2018.

\bibitem{Chen2017}
W.~Chen, X.~Chen, J.~Zhang, and K.~Huang.
\newblock Beyond triplet loss: a deep quadruplet network for person
  re-identification.
\newblock In {\em CVPR}, 2017.

\bibitem{Chen2017b}
W.~Chen, X.~Chen, J.~Zhang, and K.~Huang.
\newblock A multi-task deep network for person re-identification.
\newblock In {\em AAAI}, 2017.

\bibitem{Corbiere2017}
C.~Corbiere, H.~Ben-Younes, A.~Ram{\'e}, and C.~Ollion.
\newblock Leveraging weakly annotated data for fashion image retrieval and
  label prediction.
\newblock In {\em ICCVW}, 2017.

\bibitem{Csurka04}
G.~Csurka, C.~Dance, L.~Fan, J.~Williamowski, and C.~Bray.
\newblock Visual categorization with bags of keypoints.
\newblock In {\em ECCVW}, 2004.

\bibitem{Faghri2018}
F.~Faghri, D.~J. Fleet, J.~R. Kiros, and S.~Fidler.
\newblock {VSE++}: Improving visual-semantic embeddings with hard negatives.
\newblock In {\em BMVC}, 2018.

\bibitem{Gong2013}
Y.~Gong, S.~Lazebnik, A.~Gordo, and F.~Perronnin.
\newblock Iterative quantization: A procrustean approach to learning binary
  codes for large-scale image retrieval.
\newblock {\em IEEE TPAMI}, 2013.

\bibitem{Gordo2016}
A.~Gordo, J.~Almaz{\'a}n, J.~Revaud, and D.~Larlus.
\newblock Deep image retrieval: Learning global representations for image
  search.
\newblock In {\em ECCV}, 2016.

\bibitem{Gordo2017}
A.~Gordo, J.~Almaz\'an, J.~Revaud, and D.~Larlus.
\newblock End-to-end learning of deep visual representations for image
  retrieval.
\newblock {\em IJCV}, 2017.

\bibitem{Gordo2012}
A.~Gordo, J.~A. Rodriguez-Serrano, F.~Perronnin, and E.~Valveny.
\newblock Leveraging category-level labels for instance-level image retrieval.
\newblock In {\em CVPR}, 2012.

\bibitem{Guy2018}
I.~Guy, A.~Nus, D.~Pelleg, and I.~Szpektor.
\newblock Care to share?: Learning to rank personal photos for public sharing.
\newblock In {\em International Conference on Web Search and Data Mining},
  2018.

\bibitem{Hardwood2017}
B.~Hardwood, V.~Kumar B~G, G.~Carneiro, I.~Reid, and T.~Drummond.
\newblock Smart mining for deep metric learning.
\newblock In {\em ICCV}, 2017.

\bibitem{Harwood2017}
B.~Harwood, G.~Carneiro, I.~Reid, and T.~Drummond.
\newblock Smart mining for deep metric learning.
\newblock In {\em ICCV}, 2017.

\bibitem{Hazan2010}
T.~Hazan, J.~Keshet, and D.~A. McAllester.
\newblock Direct loss minimization for structured prediction.
\newblock In {\em NIPS}, 2010.

\bibitem{He2018b}
K.~He, F.~Cakir, S.~A. Bargal, and S.~Sclaroff.
\newblock Hashing as tie-aware learning to rank.
\newblock In {\em CVPR}, 2018.

\bibitem{He2018}
K.~He, Y.~Lu, and S.~Sclaroff.
\newblock Local descriptors optimized for average precision.
\newblock In {\em CVPR}, 2018.

\bibitem{He2016}
K.~He, X.~Zhang, S.~Ren, and J.~Sun.
\newblock Deep residual learning for image recognition.
\newblock In {\em CVPR}, 2016.

\bibitem{Henderson2016}
P.~Henderson and V.~Ferrari.
\newblock End-to-end training of object class detectors for mean average
  precision.
\newblock In {\em ACCV}, 2016.

\bibitem{Hermans2017}
A.~Hermans, L.~Beyer, and B.~Leibe.
\newblock In defense of the triplet loss for person re-identification.
\newblock {\em arXiv preprint}, 2017.

\bibitem{Jegou2012}
H.~J{\'e}gou and O.~Chum.
\newblock Negative evidences and co-occurences in image retrieval: The benefit
  of {PCA} and whitening.
\newblock In {\em ECCV}. 2012.

\bibitem{Jegou14}
H.~J\'egou and A.~Zisserman.
\newblock Triangulation embedding and democratic aggregation for image search.
\newblock In {\em CVPR}, 2014.

\bibitem{Kalantidis2016}
Y.~Kalantidis, C.~Mellina, and S.~Osindero.
\newblock Cross-dimensional weighting for aggregated deep convolutional
  features.
\newblock In {\em ECCVW}, 2016.

\bibitem{Kingma2014}
D.~P. Kingma and J.~Ba.
\newblock Adam: A method for stochastic optimization.
\newblock {\em arXiv preprint}, 2014.

\bibitem{APstudy2005}
K.~Kishida.
\newblock Property of average precision and its generalization: An examination
  of evaluation indicator for information retrieval experiments.
\newblock {\em NII Technical Reports}, 2005.

\bibitem{Liu2011}
T.~Liu.
\newblock {\em Learning to Rank for Information Retrieval}.
\newblock Springer, 2011.

\bibitem{Liu2009}
T.-Y. Liu et~al.
\newblock Learning to rank for information retrieval.
\newblock {\em Foundations and Trends in Information Retrieval}, 2009.

\bibitem{Liu2016}
Z.~Liu, P.~Luo, S.~Qiu, X.~Wang, and X.~Tang.
\newblock {DeepFashion}: Powering robust clothes recognition and retrieval with
  rich annotations.
\newblock In {\em CVPR}, 2016.

\bibitem{Lowe2004}
D.~G. Lowe.
\newblock Distinctive image features from scale-invariant keypoints.
\newblock {\em IJCV}, 2004.

\bibitem{Manmatha2017}
R.~Manmatha, C.-Y. Wu, A.~J. Smola, and P.~Kr{\"a}henb{\"u}hl.
\newblock Sampling matters in deep embedding learning.
\newblock In {\em ICCV}, 2017.

\bibitem{Mishchuk2017}
A.~Mishchuk, D.~Mishkin, F.~Radenovi\'{c}, and J.~Matas.
\newblock Working hard to know your neighbor's margins: Local descriptor
  learning loss.
\newblock In {\em NIPS}, 2017.

\bibitem{Mohapatra2014}
P.~Mohapatra, C.~Jawahar, and M.~P. Kumar.
\newblock Efficient optimization for average precision {SVM}.
\newblock In {\em NIPS}, 2014.

\bibitem{Movshovitz2017}
Y.~Movshovitz-Attias, A.~Toshev, T.~K. Leung, S.~Ioffe, and S.~Singh.
\newblock No fuss distance metric learning using proxies.
\newblock In {\em ICCV}, 2017.

\bibitem{Landmarks2017}
H.~Noh, A.~Araujo, J.~Sim, T.~Weyand, and B.~Han.
\newblock Large-scale image retrieval with attentive deep local features.
\newblock In {\em ICCV}, 2017.

\bibitem{Philbin07}
J.~Philbin, O.~Chum, M.~Isard, J.~Sivic, and A.~Zisserman.
\newblock Object retrieval with large vocabularies and fast spatial matching.
\newblock In {\em CVPR}, 2007.

\bibitem{Philbin08}
J.~Philbin, O.~Chum, M.~Isard, J.~Sivic, and A.~Zisserman.
\newblock Lost in quantization: {I}mproving particular object retrieval in
  large scale image databases.
\newblock In {\em CVPR}, 2008.

\bibitem{Raden2018}
F.~Radenovi\'{c}, A.~Iscen, G.~Tolias, Y.~Avrithis, and O.~Chum.
\newblock Revisiting {Oxford} and {Paris}: Large-scale image retrieval
  benchmarking.
\newblock In {\em CVPR}, 2018.

\bibitem{Radenovic2016}
F.~Radenovi{\'c}, G.~Tolias, and O.~Chum.
\newblock {CNN} image retrieval learns from {BoW}: Unsupervised fine-tuning
  with hard examples.
\newblock In {\em ECCV}, 2016.

\bibitem{Radenovic2018a}
F.~Radenovi{\'c}, G.~Tolias, and O.~Chum.
\newblock Fine-tuning {CNN} image retrieval with no human annotation.
\newblock {\em TPAMI}, 2018.

\bibitem{Razavian2014}
A.~S. Razavian, H.~Azizpour, J.~Sullivan, and S.~Carlsson.
\newblock {CNN} features off-the-shelf: An astounding baseline for recognition.
\newblock In {\em CVPRW}, 2014.

\bibitem{Rippel2016}
O.~Rippel, M.~Paluri, P.~Dollar, and L.~Bourdev.
\newblock Metric learning with adaptive density discrimination.
\newblock In {\em ICLR}, 2016.

\bibitem{Russakovsky2015}
O.~Russakovsky, J.~Deng, H.~Su, J.~Krause, S.~Satheesh, S.~Ma, Z.~Huang,
  A.~Karpathy, A.~Khosla, M.~Bernstein, A.~C. Berg, and L.~Fei-Fei.
\newblock {ImageNet} large scale visual recognition challenge.
\newblock {\em IJCV}, 2015.

\bibitem{Schroff2015}
F.~Schroff, D.~Kalenichenko, and J.~Philbin.
\newblock {FaceNet}: A unified embedding for face recognition and clustering.
\newblock In {\em CVPR}, 2015.

\bibitem{Shi2016}
H.~Shi, Y.~Yang, X.~Zhu, S.~Liao, Z.~Lei, W.~Zheng, and S.~Z. Li.
\newblock Embedding deep metric for person re-identification: A study against
  large variations.
\newblock In {\em ECCV}, 2016.

\bibitem{Sohn2016}
K.~Sohn.
\newblock Improved deep metric learning with multi-class $n$-pair loss
  objective.
\newblock In {\em NIPS}, 2016.

\bibitem{ProdSearch16}
H.~O. Song, Y.~Xiang, S.~Jegelka, and S.~Savarese.
\newblock Deep metric learning via lifted structured feature embedding.
\newblock In {\em CVPR}, 2016.

\bibitem{Song2016}
Y.~Song, A.~Schwing, Richard, and R.~Urtasun.
\newblock Training deep neural networks via direct loss minimization.
\newblock In {\em ICML}, 2016.

\bibitem{Spyromitros2014}
E.~Spyromitros-Xioufis, S.~Papadopoulos, I.~Y. Kompatsiaris, G.~Tsoumakas, and
  I.~Vlahavas.
\newblock A comprehensive study over {VLAD} and product quantization in
  large-scale image retrieval.
\newblock {\em IEEE Transactions on Multimedia}, 2014.

\bibitem{Tolias16}
G.~Tolias, R.~Sicre, and H.~J\'egou.
\newblock Particular object retrieval with integral max-pooling of {CNN}
  activations.
\newblock In {\em ICLR}, 2016.

\bibitem{Trotman2005}
A.~Trotman.
\newblock Learning to rank.
\newblock {\em Information Retrieval}, 2005.

\bibitem{Ustinova2016}
E.~Ustinova and V.~Lempitsky.
\newblock Learning deep embeddings with histogram loss.
\newblock In {\em NIPS}. 2016.

\bibitem{Wang2017}
C.~Wang, X.~Lan, and X.~Zhang.
\newblock How to train triplet networks with 100k identities?
\newblock In {\em ICCVW}, 2017.

\bibitem{Wang2018}
W.~Wang, Y.~Xu, J.~Shen, and S.-C. Zhu.
\newblock Attentive fashion grammar network for fashion landmark detection and
  clothing category classification.
\newblock In {\em CVPR}, 2018.

\bibitem{Yue2007}
Y.~Yue, T.~Finley, F.~Radlinski, and T.~Joachims.
\newblock A support vector method for optimizing average precision.
\newblock In {\em SIGIR}, 2007.

\end{thebibliography}
 }{\small\par}

\end{document}